\documentclass{article}
\usepackage[utf8]{inputenc}
\usepackage{authblk}
\usepackage{amsmath,amssymb,amsfonts}
\usepackage[square,numbers]{natbib}
\usepackage{graphicx}
\usepackage{tabularx}

\title{A Deep Reinforcement Learning Driving Policy for Autonomous Road Vehicles}
\author{Konstantinos Makantasis$^{1,2}$, Maria Kontorinaki$^{1,3}$, Ioannis Nikolos$^1$}
\affil{\footnotesize{$^1$School of Production Engineering and Management, Technical University of Crete, Greece\\
$^2$Institute of Digital Games, University of Malta, Malta\\
$^3$Department of Statistics and Operations Research, University of Malta, Malta}}
\date{}

\begin{document}

\maketitle

\small{\noindent \textbf{Abstract.} This work regards our preliminary investigation on the problem of path planning for autonomous vehicles that move on a freeway. We approach this problem by proposing a driving policy based on Reinforcement Learning. The proposed policy makes minimal or no assumptions about the environment, since no \textit{a priori} knowledge about the system dynamics is required. We compare the performance of the proposed policy against an optimal policy derived via Dynamic Programming and against manual driving simulated by SUMO traffic simulator.\\
\\
\textbf{Word count}: 2981}

\section{Introduction}
\label{sec:introduction}
%Building self-driving cars is an active area of research \cite{cosgun2017towards} for its high potential in leading to road networks that are safer more ore efficient. Although vehicle automation has already led to great achievements in supporting the driver in various tasks \cite{reimer2010evaluation}, rising the level of automation to fully-automated driving is an extremely challenging problem, mainly due to the complexity of real-world environments. 
According to \cite{donges1999conceptual}, autonomous driving tasks can be classified into three categories; \textit{navigation}, \textit{guidance}, and \textit{stabilization}. Navigation tasks are responsible for generating road-level routes, guidance tasks are responsible for guiding vehicles along these routes by generating tactical maneuver decisions, and stabilization tasks are responsible for translating tactical decisions into reference trajectories and then low-level controls. 

In this work, we focus on tactical level guidance, and, specifically, we aim to contribute towards the development of a robust real-time driving policy for autonomous vehicles that move on a highway. The driving policy development problem is formulated from an autonomous vehicle perspective, and, thus, there is no need to make any assumptions regarding the kind of other vehicles (manual driving or autonomous) that occupy the road. The proposed methodology approaches the problem of driving policy development by exploiting recent advances in \textit{Reinforcement Learning} (RL).

\subsection{Related Work}

The problem of path planning for autonomous vehicles can be seen as a problem of generating a sequence of states that must be tracked by the vehicle. Under certain assumptions, simplifications and conservative estimates, heuristic rules can be used towards this direction \cite{werling2008robust}. These methods, however, are often tailored for specific environments and do not generalize \cite{fletcher2008cornell} to complex real world environments and diverse driving situations. 

\textit{Optimal control} methods aim to overcome these limitations by allowing for the concurrent consideration of environment dynamics and carefully designed objective functions for modelling the goals to be achieved \cite{anderson2010optimal}. Optimal control approaches have been proposed for cooperative merging on highways \cite{ntousakis2016optimal}, for obstacle avoidance \cite{carvalho2014stochastic}, and for generating "green" trajectories \cite{typaldos2018minimization} or trajectories that maximize passengers' comfort \cite{makantasis2018motorway}. Although, optimal control methods are quite popular, there are still open issues regarding the decision making process. First, these approaches usually map the optimal control problem to a nonlinear program, the solution of which generally corresponds to a local optimum for which global optimality guarantees may not hold, and, thus, safety constraints may be violated. Second, the efficiency of these approaches is dependent on the model of the environment. In many cases, however, that model is assumed to be represented by simplified observation spaces, transition dynamics and measurements mechanisms, limiting the generality of these methods to complex scenarios. Finally, optimal control methods are not able to generalize, i.e., to associate a state of the environment with a decision without solving an optimal control problem even if exactly the same problem has been solved in the past.

Very recently, RL methods have been proposed as a challenging alternative  towards the development of driving policies. RL approaches alleviate the strong dependency on environment models and dynamics, and, at the same time, can fully exploit the recent advances in deep learning \cite{mnih2015human}. Along this line of research, RL methods have been proposed for intersection crossing and lane changing  \cite{isele2017navigating,mukadam2017tactical}, as well as, for double merging scenarios \cite{shalev2016safe}.

\subsection{Our Contribution}
We propose a RL driving policy based on the exploitation of a Double Deep Q-Network (DDQN) \cite{van2016deep}. The derived policy is able to guide an autonomous vehicle that move on a highway, and at the same time take into consideration passengers' comfort via a carefully designed objective function. To the best of our knowledge, this work is one of the first attempts that try to derive a RL policy targeting unrestricted highway environments, which are occupied by both autonomous and manual driving vehicles. Moreover, this work provides insights to the trajectory planning problem, by comparing the proposed policy against an optimal policy derived using Dynamic Programming (DP). Finally, we investigate the generalization ability and stability of the proposed RL policy using the established SUMO microscopic traffic simulator.    

\section{Problem Description and Assumptions}
\label{sec:problem_formulation}
We consider the path planning problem for an autonomous vehicle that moves on freeway, which is also occupied by manual driving vehicles. Without loss of generality, we assume that the freeway consists of three lanes. The driving policy should generate a collision-free trajectory, which should permit the autonomous vehicle to move forward with a desired speed, and, at the same time, minimize its longitudinal and lateral accelerations (passengers' comfort). The aforementioned three criteria are the objectives of the driving policy, and thus, the goal that the RL algorithm should achieve.

Moreover, we do not assume any communication between vehicles. Instead, the autonomous vehicle estimates the position and the velocity of its surrounding vehicles using sensors installed on it. The state representation of the environment, includes information that is associated solely with the position and the velocity of the vehicles. Furthermore, we assume that the freeway does not contain any turns. However, the generated vehicle trajectory essentially reflects the vehicle longitudinal position, speed, and its traveling lane, and, therefore, for the trajectory specification, possible curvatures may be aligned to form an equivalent straight section. Finally, the trajectory of the autonomous vehicle can be fully described by a sequence of high-level goals that the vehicle should achieve within a specific time interval. We assume that the mechanism which translates these goals to low-level controls and implements them is given. Based on the aforementioned problem description and underlying assumptions, the objective of this work is to derive a function that will map the information about the autonomous vehicle, as well as, its surrounding environment to a specific goal.

\section{Driving Policy}
\label{sec:driving_policy}

\subsection{The reinforcement learning framework}
\label{sec:RL-based_driving_policy}
In the RL framework, an agent interacts with the environment in a sequence of actions, observations, and rewards. At each time step $t$, the agent (in our case the autonomous vehicle) observes the state of the environment $s_t \in \mathcal S$ and it selects an action $a_t \in \mathcal A$, where $\mathcal S$ and $\mathcal A=\{1,\cdots, K\}$ are the state and action spaces. As the consequence of applying the action $a_t$ at state $s_t$, the agent receives a scalar reward signal $r_t$. The goal of the agent is to interact with the environment by selecting actions in a way that maximizes the cumulative future rewards. The interaction of the agent with the environment can be explicitly defined by a policy function $\pi:\mathcal S \rightarrow \mathcal A$ that maps states to actions. In this work we exploit a DDQN for approximating an optimal policy, i.e., an action selection strategy that maximizes cumulative future rewards. Due to space limitations we are not describing the DDQN model, we refer, however, the interested reader to \cite{van2016deep}. In order to train the DDQN, we describe, in the following, the state representation, the action space, and the design of the reward signal. 

\begin{figure}[t]
	\begin{minipage}{1.0\linewidth}
		\centering
		\centerline{\includegraphics[width=1.0\linewidth]{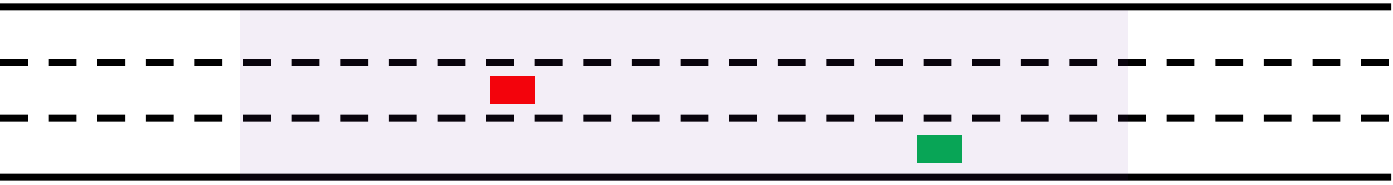}}
		\centering (a)
		\vspace{0.1in}
	\end{minipage} 
	\begin{minipage}{1.0\linewidth}
		\centering
		\centerline{\includegraphics[width=1.0\linewidth]{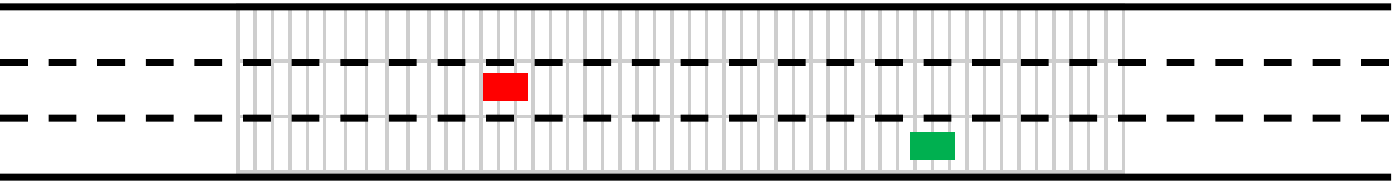}}
		\centering (b)
	\end{minipage} 
	\caption{State representation. The autonomous vehicle is represented by the red rectangle. (a) The purple shaded area corresponds to the sensed environment. (b) Discretization of sensed environment.}
	\label{fig:state}
\end{figure}

\subsection{State Representation}
\label{ssec:state}
We assume that the autonomous vehicle can sense its surrounding environment that spans 75 meters behind it and 100 meters ahead of it, as well as, its two adjacent lanes, see Fig. \ref{fig:state}(a), and it can estimate the relative positions and velocities of other vehicles that are present in these area. Note that given current LiDAR and camera sensing technologies such an assumption can be considered valid. The sensed area is discretized into tiles of one meter length, see Fig. \ref{fig:state}(b), and the value of vehicles' longitudinal velocity (including the autonomous vehicle) is assigned to the tiles beneath of them. The value of zero is given to all non occupied tiles that belong to the road, and -1 to tiles outside of the road (the autonomous vehicle can sense an area outside of the road if it occupies the left-/right-most lane). This state representation is a matrix that contains information about the absolute velocities of vehicles, as well as, relative positions of other vehicles with respect to the autonomous vehicle. The vectorized form of this matrix is used to represent the state of the environment.

\subsection{Action Representation}
The authors of \cite{liu2018elements} argue that low-level control tasks can be less effective and/or robust for tactical level guidance. For this reason we construct an action set that contains high-level actions. Specifically, we define seven available actions; i) change lane to the left or right, ii) accelerate or decelerate with a constant acceleration or deceleration of $1m/s^2$ or $2m/s^2$, and iii) move with the current speed at the current lane. For the acceleration and deceleration actions feasible acceleration and deceleration values are used. Moreover, the autonomous vehicle is making decisions by selecting one action every \textit{one second}, which implies that lane changing actions are also feasible.

\subsection{Reward Signal Design}
Designing appropriate rewards signals is the most important tool for shaping the behavior of the driving policy. The autonomous vehicle should be able to avoid collisions, move with a desired speed, and avoid unnecessary lane changes and accelerations. Therefore, the reward signal must reflect all these objectives by employing one penalty function for collision avoidance, one that penalizes deviations from the desired speed and two penalty functions for unnecessary lane changes and accelerations.

The penalty function for collision avoidance should feature high values at the gross obstacle space, and low values outside of that space. To this end, we adopt the exponential penalty function
\begin{equation}
f(\delta_i) = \begin{cases}
e^{-(\delta_i - \delta_0)} &\text{ if } l_e = l_i \\
0 &\text{ otherwise}
\end{cases},
\label{eq:distance_penalty}
\end{equation}
where $\delta_i$ is the longitudinal distance between the autonomous vehicle and the $i$-th obstacle, $\delta_0$ stands for the minimum safe distance, and, $l_e$ and $l_i$ denote the lanes occupied by the autonomous vehicle and the $i$-th obstacle. If the value of (\ref{eq:distance_penalty}) becomes greater or equal to one, then the driving situation is considered very dangerous and it is treated as a collision. 

The vehicle mission is to advance with a longitudinal speed close to a desired one. Thus, the quadratic term 
\begin{equation}
\label{eq:speed_penalty}	
h(v) = (v - v_d)^2
\end{equation}
that penalizes the deviation between real vehicles speed and its desired speed is used. Variable $v$ and $v_d$ stand for the real and the desired speed of the autonomous vehicle. 

We also introduce two penalty terms for minimizing accelerations and lane changes. For penalizing accelerations we use the term
\begin{equation}
    a(v_t, v_{t-1}) = (v_t - v_{t-1})^2,
\end{equation}
and for penalizing lane changes the term
\begin{equation}
    g(l_t, l_{t-1}) = \mathbb I(l_{e,t}\neq l_{e,t-1}).
\end{equation}
Variables $v_k$ and $l_k$ correspond to the speed and lane of the autonomous vehicle at time step $k$, while $\mathbb I(\cdot$) is the indicator function. The total rewards at time step $t$ is the negative weighted sum of the aforementioned penalties:
\begin{equation}
%\begin{split}
r_t = - w_1 \sum_{i=1}^{O_t}f_t(\delta_i) - w_2 h_t(v_t) - w_3\sum_{i=1}^{O_t} \mathbb I(f_t(\delta_i) \geq 1) - w_4 a(v_t,v_{t-1}) - w_5 g(l_t, l_{t-1})
%\end{split}
\label{eq:reward}
\end{equation}
In (\ref{eq:reward}) the third term penalizes collisions and variable $O_t$ corresponds to the total number of obstacles that can be sensed by the autonomous vehicle at time step $t$. The selection of weights defines the importance of each penalty function to the overall reward. In this work the weights were set, using a trial and error procedure, as follows: $w_1=1$, $w_2=0.5$, $w_3=20$, $w_4=0.01$, $w_5=0.01$.

Before proceeding to the experimental results, we have to mention that the employed DDQN comprises of two identical neural networks with two hidden layers with 256 and 128 neurons. Also, the synchronization between the two neural networks, see \cite{van2016deep}, is realized every 1000 epochs.

\section{Experimental Results}
Two different sets of experiments were conducted. In the first set of experiments, we developed and utilized a simplified custom made microscopic traffic simulator, while, the second set employs the established SUMO microscopic traffic simulator. 

\subsection{First set of experiments}
The custom made simulator moves the manual driving vehicles with constant longitudinal velocity using the kinematics equations. Moreover, the manual driving vehicles are not allowed to change lanes. Despite its simplifying setting, this set of experiments allow us to compare the RL driving policy against an optimal policy derived via DP. 

For training the DDQN, driving scenarios of 60 seconds length were generated. In these scenarios one vehicle enters the road every two seconds, while the tenth vehicle that enters the road is the autonomous one. All vehicles enter the road at a random lane, and their initial longitudinal velocity was randomly selected from a uniform distribution ranging from 12m/s to 17m/s. Finally, the desired speed of the autonomous vehicle was set equal to 21m/s.

We compared the RL driving policy against an optimal policy derived via DP under four different road density values. For each one of the different densities 100 scenarios of 60 seconds length were simulated. In these scenarios, the simulator moves the manual driving vehicles, while the autonomous vehicle moves by following the RL policy and by solving a DP problem (which utilizes the same objective functions and actions as the RL algorithm). Finally, we extracted statistics regarding the number of collisions and lane changes, and the percentage of time that the autonomous vehicle moves with its desired speed for both the RL and DP policies. At this point it has to be mentioned that DP is not able to produce the solution in real time, and it is just used for benchmarking and comparison purposes. %On the contrary the RL policy, at a given state can select an action very fast, since this selection corresponds to one evaluation of the neural network function at the given state. 

\begin{table}[t]
	\centering
	\caption{Driving behavior evaluation of the RL and DP driving policies, in terms of total number of collision and lane changes for 100 scenarios and percentage of time that the vehicle moves with its desired speed. }
	\newcolumntype{L}[1]{>{\hsize=#1\hsize\raggedright\arraybackslash}X}%
	\newcolumntype{C}[1]{>{\hsize=#1\hsize\centering\arraybackslash}X}%
	\label{table:1}
	
	\begin{tabularx}{1.0\linewidth}{L{8.5}C{6.0}C{7.0}C{9.0}}
		\hline \hline 
		\textbf{1 veh./8 sec.} & Collisions & Lane changes & Desired speed (\%) \\ \hline
		DP policy   & 0          & 84           & 85 \\ \hline
		RL policy   & 0          & 81           & 73    \\ 
		\hline
        \hline
		\textbf{1 veh./4 sec.} &  &   &  \\ \hline
		DP policy   & 0          & 127          & 83 \\ \hline
		RL policy   & 0          & 115          & 64    \\ 
		\hline
        \hline
		\textbf{1 veh./2 sec.} &  & &  \\ \hline
		DP policy   & 0          & 120          & 87 \\ \hline
		RL policy   & 0          & 108          & 62    \\ 
		\hline
        \hline
		\textbf{1 veh./1 sec.} &  &  &  \\ \hline
		DP policy   & 0          & 70           & 72 \\ \hline
		RL policy   & 2          & 62           & 56    \\ \hline \hline

	\end{tabularx}
\end{table}

Table \ref{table:1} summarizes the results of this comparison. The four different densities are determined by the rate at which the vehicles enter the road, that is, 1 vehicle enters the road every 8, 4, 2, and 1 seconds. The RL policy is able to generate collision free trajectories, when the density is less than or equal to the density used to train the network. However, for larger density the RL policy produced 2 collisions in 100 scenarios. In terms of efficiency, the optimal DP policy is able to perform more lane changes and advance the vehicle faster.  

\begin{table}[t]
	\centering
	\caption{Total number of collisions during 100 scenarios, when different magnitudes of measurement errors are introduced.}
	\newcolumntype{L}[1]{>{\hsize=#1\hsize\raggedright\arraybackslash}X}%
	\newcolumntype{C}[1]{>{\hsize=#1\hsize\centering\arraybackslash}X}%
	\label{table:2}
	
	\begin{tabularx}{1.0\linewidth}{L{8.5}C{6.0}C{7.0}C{9.0}}
		\hline \hline 
	 Measurement error & $\pm$5\% & $\pm$10\% & $\pm$15\% \\ \hline
		1 veh./8 sec.   & 0          & 0           & 0 \\ \hline
		1 veh./4 sec.   & 0          & 0           & 0 \\ \hline
		1 veh./2 sec.   & 0          & 1           & 1 \\ \hline
		1 veh./1 sec.   & 3          & 4           & 4 \\ \hline \hline 
		
	\end{tabularx}
\end{table}

We also evaluated the robustness of the RL policy to measurement errors regarding the position of the manual driving vehicles. At each time step, measurement errors proportional to the distance between the autonomous vehicle and the manual driving vehicles are introduced. We used three different error magnitudes; $\pm5\%$, $\pm10\%$, and $\pm15\%$. The RL policy was evaluated in terms of collisions in 100 driving scenarios of 60 seconds length for each error magnitude. When the density value is less than the density used to train the network the RL policy is very robust to measurement errors and produces collision free trajectories, see Table \ref{table:2}. When the density is equal to the one used for training, the RL policy can produce collision free trajectories only for small measurement errors, while for larger errors it produced 1 collision in 100 driving scenarios. Finally, when the density becomes larger, the performance of the RL policy deteriorates.

\subsection{Second set of experiments}
In the second set of experiments we evaluate the behavior of the autonomous vehicle when it follows the RL policy and when it is controlled by SUMO. We trained the RL policy using scenarios generated by the SUMO simulator. During the generation of scenarios, all SUMO safety mechanisms are enabled for the manual driving vehicles and disabled for the autonomous vehicle. Furthermore, we do not permit the manual driving cars to implement cooperative and strategic lane changes. Such a configuration for the lane changing behavior, impels the autonomous vehicle to implement maneuvers in order to achieve its objectives. Moreover, in order to simulate realistic scenarios two different types of manual driving vehicles are used; vehicles that want to advance faster than the autonomous vehicle and vehicles that want to advance slower. Finally, the density was equal to 600 veh/lane/hour. 

For the evaluation of the trained RL policy, we simulated i) 100 driving scenarios during which the autonomous vehicle follows the RL driving policy, ii) 100 driving scenarios during which the default configuration of SUMO was used to move forward the autonomous vehicle, and iii) 100 scenarios during which the behavior of the autonomous vehicle is the same as the manual driving vehicles, i.e. it does not perform strategic and cooperative lane changes. The duration of all simulated scenarios was 60 seconds. We simulated scenarios for two different driving conditions. In the first one the desired speed for the slow manual driving vehicles was set to $18m/s$, while in the second one to $16m/s$. For both driving conditions the desired speed for the fast manual driving vehicles was set to $25m/s$. Furthermore, in order to investigate how the presence of uncertainties affects the behavior of the autonomous vehicle, we simulated scenarios where drivers' imperfection was introduced by appropriately setting the $\sigma$ parameter in SUMO. Finally, the behavior of the autonomous vehicles was evaluated in terms of i) collision rate, ii) average lane changes per scenario, and iii) average speed per scenario.  

\begin{table}[t]
	\centering
	\caption{Driving behavior evaluation. \textit{SUMO default} corresponds to the default SUMO configuration, while \textit{SUMO manual} to the case where the behavior of the autonomous vehicle is the same as the  manual driving vehicles.}
	\newcolumntype{L}[1]{>{\hsize=#1\hsize\raggedright\arraybackslash}X}%
	\newcolumntype{C}[1]{>{\hsize=#1\hsize\centering\arraybackslash}X}%
	\label{table:3}
	
	\begin{tabularx}{1.0\linewidth}{L{10.9}C{4.7}C{6.7}C{7.7}}
		\hline \hline 
		\multicolumn{4}{ c }{\textbf{Desired speed for slow vehicles 18m/s}}\\ \hline
		& Collisions & Lane changes & Average speed \\ \hline
		
		RL policy ($\sigma=0.0$) & 2\% & 1.93 & 20.71 \\ \hline
		
		SUMO default ($\sigma=0.0$)    & 0\% & 0.99 & 20.22    \\ \hline
		
		SUMO manual ($\sigma=0.0$)     & 0\% & 0.0  & 19.48    \\ \hline
		
		RL policy ($\sigma=0.5$) & 3\% & 1.92 & 20.09 \\ \hline
		
		SUMO default ($\sigma=0.5$)    & 0\% & 0.89 & 19.57    \\ \hline
		
		SUMO manual ($\sigma=0.5$)     & 0\% & 0.0  & 19.05    \\ \hline \hline 
		\multicolumn{4}{ c }{\textbf{Desired speed for slow vehicles 16m/s}}\\ \hline
		& Collisions & Lane changes & Average speed \\ \hline
		RL policy ($\sigma=0.0$) & 2\% & 2.02 & 20.04 \\ \hline
		
		SUMO default ($\sigma=0.0$)    & 0\% & 0.33 & 18.41    \\ \hline
		
		SUMO manual ($\sigma=0.0$)     & 0\% & 0.0  & 17.47    \\ \hline
		
		RL policy ($\sigma=0.5$) & 4\% & 1.83 & 19.87 \\ \hline
		
		SUMO default ($\sigma=0.5$)    & 0\% & 0.31 & 17.67    \\ \hline
		
		SUMO manual ($\sigma=0.5$)     & 0\% & 0.0  & 17.26    \\ \hline \hline \\

	\end{tabularx}
\end{table}

In Table \ref{table:3}, \textit{SUMO default} corresponds to the default SUMO configuration for moving forward the autonomous vehicle, while \textit{SUMO manual} to the case where the behavior of the autonomous vehicle is the same as the manual driving vehicles. Irrespective of whether a perfect ($\sigma=0$) or an imperfect ($\sigma=0.5$) driver is considered for the manual driving vehicles, the RL policy is able to move forward the autonomous vehicle faster than the SUMO simulator, especially when slow vehicles are much slower than the autonomous one. In order to achieve this, RL policy implements more lane changes per scenario. However, it results to a collision rate of 2\%-4\%, which is its main drawback. No guarantees for collision-free trajectory is the price paid for deriving a learning based approach capable of generalizing to unknown driving situations and inferring with minimal computational cost, driving actions. Although this drawback is prohibitive for applying such a policy in real world environments, a mechanism can be developed to translate the actions proposed by the RL policy in low level controls and then implement them in a safe aware manner. The development of such a mechanism is the topic of our ongoing work, which comes to extend this preliminary study and provide a complete methodology for deriving RL collision-free policies.

\section{Conclusions}
\label{sec:5}
In this work, we employed the DDQN model to derive a RL driving policy for an autonomous vehicle that moves on a highway. The proposed policy makes no assumptions about the environment, it does not require any knowledge about the system dynamics. Moreover, it is able to produce actions with very low computational cost via the evaluation of a function, and what is more important, it is capable of generalizing to previously unseen driving situations. The derived driving policy, however, it cannot guarantee a collision free trajectory. For this reason, there is an imminent need for developing a low-level mechanism capable to translate the action coming from the RL policy to low-level commands, and,  then implement them in a safe aware manner. The development of such a mechanism is the main objective of our ongoing work. 

\section{Acknowledgement}
This research is implemented through and has been financed by the Operational Program ''Human Resources Development, Education and Lifelong Learning'' and is co-financed by the European Union (European Social Fund) and Greek national funds.

\bibliographystyle{abbrvnat}
\bibliography{refs}
\end{document}